\documentclass[]{svjour3}
\usepackage{amsmath}
\usepackage{amssymb}
\usepackage{amsfonts}
\usepackage{graphicx}
\usepackage[]{float}
\usepackage{multirow}
\usepackage{array}
\usepackage{url}
\usepackage{natbib}

\newcommand{\textfigure}[1]{{\parbox{0.8\textwidth}{
\fontsize{9pt}{9pt}
\selectfont
\fontdimen2\font=0.3em
\fontdimen3\font=0.1em
\fontdimen4\font=0.1em
\fontdimen7\font=0.1em
\hyphenchar\font=`\-
#1}}}
\newcolumntype{H}{>{\setbox0=\hbox\bgroup}c<{\egroup}@{}}
\newcommand{\EncSum}{EncSum}

\begin{document}
\title{Encoded Summarization: Summarizing Documents into Continuous Vector Space for Legal Case Retrieval}
\titlerunning{Encoded Summarization for Legal Case Retrieval}
\author{Vu Tran$\dagger$, Minh Le Nguyen$\dagger$, Satoshi Tojo$\dagger$, and Ken Satoh$\ddagger$}
\authorrunning{Tran et al.}
\institute{$\dagger$Japan Advanced Institute of Science and Technology; $\ddagger$The National Institute of Informatics, Japan}
\journalname{AI and Law}
\date{Published 2020-01-25 in AI and Law. DOI: \url{https://doi.org/10.1007/s10506-020-09262-4}}
\def\makeheadbox{}
\maketitle

\begin{abstract}
We present our method for tackling a legal case retrieval task by introducing our method of encoding documents by summarizing them into continuous vector space via our phrase scoring framework utilizing deep neural networks. On the other hand, we explore the benefits from combining lexical features and latent features generated with neural networks. Our experiments show that lexical features and latent features generated with neural networks complement each other to improve the retrieval system performance. Furthermore, our experimental results suggest the importance of case summarization in different aspects: using provided summaries and performing encoded summarization. Our approach achieved F1 of 65.6\% and 57.6\% on the experimental datasets of legal case retrieval tasks. 
\end{abstract}

\section{Introduction}

Automatic legal document processing systems can speed up significantly the work of experts, which, otherwise, requires significant time and efforts. One crucial kind of such systems, automatic information retrieval whose systems, in place of experts, process over enormous amount of documents, for example, legal case reports, which are accumulated rapidly over time (the number of filings in the U.S. district courts for civil cases and criminal defendants is 344,787 in 2017 \footnote{http://www.uscourts.gov/statistics-reports/judicial-business-2017}). A case document contains a large volume of contents as the case  may  last  days  or even years. This one problem challenges the construction of an effective automatic legal case retrieval system.

In 2018, the Competition on Legal Information Extraction and Entailment (COLIEE) introduced two new tasks involving processing legal case documents: a legal case retrieval task and a legal case entailment task together with the previously introduced other two tasks about statue law~\cite{COLIEE2018}. The legal case retrieval task is ``to explore and evaluate case law retrieval technologies that are both effective and reliable". The legal case entailment task is to ``identify which paragraph in the noticed case entails the decision" given the noticed cases (assumed to be correctly retrieved) and a decision. A legal case document processing system fulfilling the two tasks would benefit lawyers in finding relevant information to construct arguments for their own objectives. This work tackles the first task: legal case retrieval.

In the format of COLIEE 2018 and COLIEE 2019, the legal case retrieval task involves reading a new case $q$, and extracting supporting cases $c^*_1$, $c^*_2$, ..., $c^*_n$ for the decision of $q$ from a given list of candidate cases. The candidate cases that support for the decision of a new case are called 'noticed cases'. 

We tackle the task of finding the cases having supporting relationship with a new case indirectly through similarity measure. Our system extracts various kinds of features indicating the similarity of a new case and a previous case. The system is trained to score the relevance of the two cases by weighting the extracted similarity features. The similarity features are computed by comparing the different kinds of representations of the two cases including textual and vector representations. While the system does not provide definite answer to the supporting relationship, it learns from data to predict relevant cases by learning how the similarity features and the supporting relationship are related.

\begin{table}
    \centering
    \caption{Statistics of candidate case documents in COLIEE 2018 and 2019 training data. (*) Only count documents having an expert summary.}
    \begin{tabular}{|l|r|r|r|r|}
    \hline \hline
        & \multicolumn{2}{c}{2018} & \multicolumn{2}{|c|}{2019} \\
        \textbf{Property}  & \textbf{Max} & \textbf{Avg.}  & \textbf{Max} & \textbf{Avg.} \\
        \hline \hline
        \#words/doc & 85,551 & 5,690 & 9,666 & 2,665 \\
        \#paragraphs/doc & 1,117 & 43 & 119 & 22 \\
        \#summary-words/doc* & 8,827 & 589 & 3,085 & 242 \\
        \hline \hline
    \end{tabular}
    \label{tbl:data-stat}
\end{table}

\begin{figure}
    \centering
    \textfigure{
\textbf{Summary:}

A human rights complaint alleged the federal government's under-funding of welfare services for on-reserve First Nations children resulted in a lower level of services for those children than for other Canadian children whose welfare services were provincially funded. /*~...~*/

The Federal Court held that, while the Tribunal had the power to decide this issue in advance of a full hearing on the merits, the process followed was not fair. /*~...~*/

Administrative Law - Topic 547

The hearing and decision - Decisions of the tribunal - Reasons for decision - When required - [See second Civil Rights - Topic 7046].

Administrative Law - Topic 2608

Natural justice - Evidence and proof - Extraneous or irrelevant considerations - [See first Civil Rights - Topic 7046].

/*~...~*/

\textbf{Paragraphs:}

\textbf{[1]} Mactavish, J. : The Government of Canada funds child welfare services for First Nations children living on reserves. The provinces fund child welfare services for all other Aboriginal and non-Aboriginal children.

\textbf{[2]} The First Nations Child and Family Caring Society and the Assembly of First Nations filed a human rights complaint with the Canadian Human Rights Commission in which they allege that the Government of Canada under-funds child welfare services for on-reserve First Nations children.~/*~...~*/

/*~...~*/

\textbf{[254]} In my view, the ordinary meaning of the phrase ``differentiate adversely in relation to any individual" on a prohibited ground of discrimination is to treat someone differently than you might otherwise have done because of the individual's membership in a protected group.~/*~...~*/
 
/*~...~*/

\textbf{[395]} As a result, the three applications for judicial review are granted. /*~...~*/

\textbf{[396]} THIS COURT ORDERS AND ADJUDGES that /*~...~*/
}

    \caption{Illustration of a legal case document from Federal Court of Canada. ``/*~...~*/": omitted contents. Other information about citing cases, noticed cases, notices statutes, etc. are omitted.}
    \label{fig:case-example}
\end{figure}

A legal case document contains case details and may contain other information such as citing cases, noticed cases, notices statutes, or editor drafted summaries. The case details are presented in form of paragraphs which can be fact statements, discussed legal points or the case decision (Fig.~\ref{fig:case-example}). The summary, if present, contains court decision, decisive facts, decisive legal points, and several key phrases, which is drafted by an editor.

Legal case documents usually contain huge amount of contents. As in Table~\ref{tbl:data-stat}, in COLIEE 2018, a legal case document contains $\approx$5.7K words and 43 paragraphs in average, and could goes over 80K words and 1K paragraphs. This challenges the efficiency of not only human experts but also automatic retrieval systems. Editor summarization condensates contents by $\approx$90\% which results in  $\approx$10\% key contents.

In COLIEE 2019, we observed the similar and different challenges. First, the candidate cases are $\approx$2.7K-token long in average (Table~\ref{tbl:data-stat}). The difficulty of reading too long texts still emerges. We may pursue the idea that using summary as the main source of information.
However, the dataset of COLIEE 2019 is different from the one of COLIEE 2018. While in COLIEE 2018, most of the candidate cases have a summary, in COLIEE 2019, more than $\approx$47K in a total of 57K candidate cases are confirmed to have no summary (indicated with the note ``This case is unedited, therefore contains no summary"). This means that summarization over candidate case requires additional effort so that we can compare a query's summary with a candidate's summary.

We develop our system with representing a legal case document from its highlight contents. One way is to look at the editor drafted summary or catchphrases if these are available. The summary concisely states the decision of the case with the main arguments supporting the decision. While ``catchphrases have an indicative function rather than informative, they present all the legal point considered instead that just summarizing the key points of a decision'' \citep{Galgani:2012:TAG:2238696.2238737}. 
Catchphrases give a quick impression on what the case is about: ``the function of catchwords is to give a summary classification of the matters dealt with in a case. [...] Their purpose is to tell the researcher whether there is likely to be anything in the case relevant to the research topic" \citep{olsson1999guide}. On one hand, catchphrases help lawyers/researchers quickly grasp the points of a case, without having to read the entire document, which saves significant time and effort for finding/studying relevant cases. On the other hand, catchphrases help improves the performance of automatic case retrieval systems.

\begin{table}
\centering
\caption{Example of catchphrases found in legal case reports.}
\label{tbl:eg-catchphrases}
\begin{tabular}{|p{0.9\textwidth}|}
\hline
MIGRATION - partner visa - appellant sought to prove domestic violence by the provision of statutory
declarations made under State legislation - ``statutory declaration" defined by the Migration
Regulations 1994 (Cth) to mean a declaration ``under" the Statutory Declarations Act 1959 (Cth)
in Div 1.5 - contrary intention in reg 1.21 as to the inclusion of State declarations under s 27 of
the Acts Interpretation Act - statutory declaration made under State legislation is not a statutory
declaration ``under" the Commonwealth Act - appeal dismissed \\
\hline

\end{tabular}
\end{table}

Despite of the benefits, catchphrases are not always available in legal case documents, and are drafted by legal experts, which requires huge efforts when considering the enormous number of legal case documents. 
It is, therefore, beneficial to build  automatic catchphrase generation systems for both old documents not having drafted catchphrases and new documents. Developing such systems, however, is challenging as  the complexity of catchphrases shown in Table \ref{tbl:eg-catchphrases}. 

Approaches for generating catchphrases are based on phrase scoring derived from common model for retrieval: lexical matching with term frequency-inverse document frequency \citep{galgani2012citation,Galgani:2012:TAG:2238696.2238737,Mandal:2017:ACI:3132847.3133102}. The approaches are bounded by the limit of lexical matching, and corpus-wide statistical information. The limit of lexical matching can be lifted by moving to distributed vector space, for instance, distributed word embeddings in which common models are \textit{Word2Vec} \citep{mikolov2013distributed} and \textit{GloVe} \citep{pennington2014glove}. Corpus-wide statistical information has limit capability to identify catchphrases which are not really specific to some document but commonly used in several others. 

In the COLIEE datasets, the legal documents may or may not have a drafted summary. Even using the drafted summary only may still result in limited performance as  we observed in the datasets that the summary of a query case may not be similar to some of its noticed cases. We would like to build a system that is able to extract more informative features or key contents from a legal case document than just the summary. 

We present our work on developing a legal case summarization system and on top of its core component - phrase scoring framework, building a legal case retrieval system.

First, we build a learning model to extract catchphrases for new documents with the knowledge from previously seen documents and the expert drafted catchphrases thereof. Our system utilizes deep neural networks which have been widely used in natural language processing \citep{doi:10.1162COLIr00312} to learn the direct relationship between gold catchphrases and document phrases. This results in our phrase scoring framework which is used to identify important phrases from a given legal case document.

On top of the phrase scoring framework, we develop our legal case document representation method which summarizes the document into continuous vector space. The representation is used as latent features for constructing case relevance ranking model, the core component of the retrieval system. 

We also explore the benefits of employing various types of similarity measurement belonging to lexical similarity (keyword matching) and semantic similarity (meaning matching).

On one hand, the lexical similarity and semantic similarity differ from each other and can potentially complement each other as well. 
The lexical similarity is obtained with approaches where the texts are compared by the direct surface forms with probably some transformations such as stemming, lemmatization, stopword removal, etc. High lexical similarity can present high matching, but low lexical similarity does not say much.

On the other hand, semantic similarity can provide the measurement where the surface forms are mismatched, for example, by paraphrasing. Semantic similarity can be learned in unsupervised fashion where common approaches are using statistical methods and benefits from huge available corpora (e.g. Wikipedia, GoogleNews, etc.) \citep{le2014distributed,levy2014dependency,mikolov2013distributed,pennington2014glove}. Those methods treat a document as bag/sequence of words equally. Other information in the documents such as important words or phrases, or the document hierarchy when considered may provide significant information.

\section{Encoded Summarization: Composing Document Vector from Phrase Scoring via Summary}\label{sec:enc-sum}
In this section, we describe the method to compose document representations from phrase scoring via summary. 
When dealing with the legal case retrieval task, we observed several obstacles. First, the candidate cases are 5.7K-token long in average. This poses the problem of understanding the reason of selecting the cases as supporting cases. We, then, chose another approach which is comparing the summaries of each query and its candidate cases. We, however, found that the summary of the query is not necessarily lexically similar to the summary of the candidate cases. Moreover, some candidate cases do not have summary at all. We would like to obtain the summary for each and every candidate cases, and furthermore, the summary should be comparable with the summary of the corresponding query. One approach is to map the summaries into vector space with word embeddings (\textit{word2vec} or \textit{GloVe}) or document-embeddings (\textit{doc2vec}). We come with another approach of document-embeddings which is to weight the document contents and perform weighted composition. For weighting the document contents, we build our phrase scoring framework to learn a scoring model based on the document summary. 

\subsection{The phrase scoring model}\label{sec:phase-1-scoring}
In this phase, we present our scoring model and how to train it using documents and their corresponding drafted summary.
\subsubsection{Constructing our scoring model architecture} 
\begin{figure}
\centering
\includegraphics[width=0.5\textwidth,trim=0 8cm 26cm 0,clip]{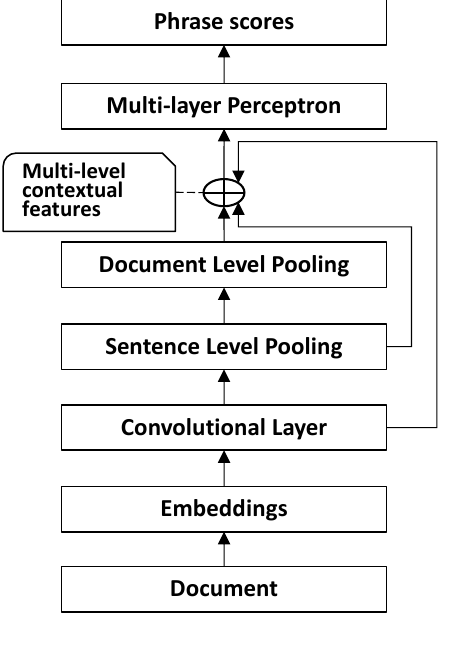}
\caption{Scoring model pipeline.}
\label{fig:arch}
\end{figure}

We score each phrase in a document based on its contexts: its words, enclosing sentence, and document. Our approach takes advantage of the core property of word embedding techniques by Google word2vec, GloVe, etc.: contextual similarity, the similarity of two words is measured as the amount of common contexts where they appear. The phrase scoring model architecture is illustrated in Fig.~\ref{fig:arch}.

We adapt convolutional neural networks (CNNs), which are successfully used in text modeling \citep{kim:2014:EMNLP2014,severyn2015learning,johnson-zhang:2015:NAACL-HLT,kalchbrenner-grefenstette-blunsom:2014:P14-1}, to encode each local context into latent feature space. Specifically, document phrase (summary phrase) features are captured by applying convolutional operations with window size $2k+1$ covering the word, $k$ left and $k$ right neighbors. 

Given a document, we denote $w^{s_i}_{j}$ as word $j^{th}$ of  sentence $i^{th}$. The features of an n-gram phrase $p_{j}=\{w_{j}, w_{j+1}, ... ,w_{j+l-1}\}$ of a sentence are captured using convolutional neural layer as follows:   

\begin{equation}
\mathbf{f_{p_{j}}}=ReLU\left(\mathbf{W}^c  \left[ \begin{array}{l}
\mathbf{v}(w_{{j}}) \\
\mathbf{v}(w_{j+1}) \\
... \\
\mathbf{v}(w_{{j+l-1}}) \\

\end{array} 
\right]\right)
\end{equation}
where, 
$\mathbf{v}(\cdot):\ \mapsto \mathbb{R}^d$: word embedding vector lookup map,
$l$: corresponding to the window size containing $l$ contiguous words,
$[\cdot] \in \mathbb{R}^{dl}$: concatenated embedding vector,
$\mathbf{W}^c\in \mathbb{R}^{c\times dl}$: convolution kernel matrix with $c$ filters,
$\mathbf{f}_{p_{j}} \in \mathbb{R}^{c} $: phrase feature vector,
$ReLU$: rectified linear unit activation. 

Sentence (catchphrase) features are, then, captured by applying max pooling over the whole sentence (catchphrase). 

\begin{equation}
\mathbf{f_{s_{i}}} = \mbox{max-pooling}_j({\mathbf{f_{p^{s_i}_{j}}}})
\end{equation}
\begin{equation}
\mathbf{f_{c_{i}}} = \mbox{max-pooling}_j({\mathbf{f_{p^{c_i}_{j}}}})
\end{equation}
where max pooling are operated over each dimension of vectors ${\mathbf{f_{p^s_{i,j}}}}$ (${\mathbf{f_{p^c_{i,j}}}}$).

Document features are captured by applying max pooling over the document (not including summary). With the same max pooling operation as above, we compute document features as: 

\begin{equation}
\mathbf{f_d} = \mbox{max-pooling}_i(\mathbf{f_{s_{i}}})
\end{equation}

The document features depend on only the document sentence, thereby, independent from the gold summary which are obviously not available for new documents.  

Finally, we apply a multilayer perceptron (MLP) with one hidden and one output layer
\begin{equation}
MLP(\mathbf{x}) = \mbox{sigmoid}(\mathbf{W}_2 \cdot \tanh(\mathbf{W}_1 \cdot \mathbf{x} + \mathbf{b}_1 ) + \mathbf{b}_2)
\end{equation}
 to compute the score of each phrase $p^{s_i}_{j}$ ($p^{c_i}_{j}$) as
\begin{equation}
P(p_s,s,d) = MLP\left(\left[ \begin{array}{l} 
{f_{p^{s_i}_{j}}} \\
{f_{s_{i}}} \\
{f_d} \\
\end{array}\right]\right)
\end{equation}
\begin{equation}
P(p_c,c,d) = MLP\left(\left[ \begin{array}{l} 
{f_{p^{c_i}_{j}}} \\
{f_{c_{i}}} \\
{f_d} \\
\end{array}\right]\right)
\end{equation}
where the hidden layer computes the phrase representative features respecting to its local use, its enclosing sentence, and its document. The word representative features are feed to the output layer to compute word score (ranging from 0.0 to 1.0).

\subsubsection{Training our scoring model} Main objective: given a document, summary phrases are ``expected" to have higher score than document phrases.  

First, we denote mean $E$ and standard deviation $std$ of word scores $P$ for each document $d$ in the following equations, which we will use to describe our objective as set of constraints, then formulated into loss function to be optimized.

\begin{equation}
E_c=E[P(p_c,c,d)]  \mbox{ where } p_c \in c, c\in d
\end{equation}

\begin{equation}
std_c=std[P(p_c,c,d)]  \mbox{ where } p_c \in c, c\in d
\end{equation}

\begin{equation}
E_s=E[P(p_s,s,d)]  \mbox{ where } p_s \in s, s\in d
\end{equation}

\begin{equation}
std_s=std[P(p_s,s,d)]  \mbox{ where } p_s \in s, s\in d
\end{equation}

\begin{equation}
E_{c,d'}=E[P(p_c,c,d')] \mbox{ where } p_c \in c, c\not\in d' 
\end{equation}
Where $p, c, s, d $ stand for phrase, summary sentence, document sentence, and the whole document respectively. $c\not\in d'$ means $c$ is not a summary of document $d'$. 

The main objective is realized by comparing the mean scores of summary phrases and document phrases:

\begin{description}
\item[(o1)] The mean score of summary phrases is higher than the mean score of document phrases: $E_c > E_s$.  
\end{description}

\begin{description}
\item[(o2)] The mean score of summary phrases is lower than document phrases when comparing a summary with a document that the summary does not belong to: $E_{c,d'} < E_{s'}$. This is the negative constraint as opposed to the constraint \textbf{o1}.
\end{description}

The above two constraints are straightforward as the positive and negative factors of the objective. However, the comparison of the mean values does not guarantee to obtain to good scoring model as the score boundaries are not considered yet.

\begin{description}
\item[(o3)] The maximum score of summary phrases is higher than the maximum score of document phrases. It is expected that there exist concise summary phrases which is typical and representative for the document but could not found in the document. Such summary phrases should get higher scores than document phrases. The estimation $E+std$ is used for representing max instead of hard max, whereby the constraint is realized as $(E_c + std_c)  >  (E_s + std_s)$.
\item[(o4)] The minimum score of summary phrases is higher than the mean score of document phrases. Once again, to emphasize the importance of summary phrases, all summary phrases should get higher score than the average score of document phrases. The estimation $E-std$ is used for representing min instead of hard min, whereby the constraint is realized as $(E_c - std_c)  >  E_s$.
\end{description}

We also add the following additional constraint to keep the scores from collapsing, which acts as regularization. 

\begin{description}
\item[(o5)] Scores should not have small variance: $std_c \not\approx 0, std_s \not\approx 0$. 
\end{description}

The loss function, hence, is composed from the constraints \textbf{(o1-5)} as follows. 

\begin{equation}
\begin{split}
\mathfrak{L}=\sum_d \max(0,m - ( & a_1(E_c - E_s)   \\
+ & a_2(\frac{1}{|\{d'\}|}\sum_{d'\neq d}{E_{s'} - E_{c,d'}})   \\
+ & b_1({(E_c + std_c)  - (E_s + std_s)})  \\
+ & b_2({(E_c - std_c)  - E_s})  \\
- & b_3(std_c) - b_4(std_s)  \\
))
\end{split}
\end{equation}

Note that rather imposing hard constraints, we compose the loss function with soft constraints. This means that some constraints may not be strictly satisfied after the training process. However, the violations of such constraints still incur certain losses and benefit the learning process. 

\subsection{Document Vector Composition}
We present our method of composing document vectors from the phrase scoring model.

Given a document, we obtain its phrase scores and internal representations at three levels: phrase level, sentence level and document level. Then, we compose the document vector as:
\begin{equation}\label{eq:enc-summary-com}
    \mathbf{g}(d) = \frac{ \sum_{i,j}{P\left(p_{j}^{s_i},s_i,d\right) \times \left[ \mathbf{f}_d ; \mathbf{f}_{s_i} ; \mathbf{f}_{p_j^{s_i}}\right]}}{ \sum_{i,j}{P\left(p_{j}^{s_i},s_i,d\right)} }
\end{equation}

Given a document, the composition weights the document contents based on their scores obtained from the phrase scoring framework. Important contents should have high contribution or affection to the final document vector. 
The component representations are multi-level contextual features which are the internal representations of the phrase scoring model. These internal representations contain the features which are learned to be used as base for scoring the surface contents. By using the multi-level contexts, the final document vector embeds the weighted multi-level contextual information including phrase level and sentence level contexts. 

This composition resembles summarization where we weight the document internal representations by its summary. Thus, we call this composition encoded summarization.

\subsection{Generating Text Summary}\label{sec:gen-text-sum}
In this phase, we generate a summary for given a document by selecting and joining document phrases scored by the phrase scoring model. The process is as follows.
\begin{itemize}
    \item Rank document phrases by their phrasal scores.
    \item Select phrases with scores from high to low.
    \item Join overlapping phrases into a longer phrase.
    \item Stop when the summary length exceeds length-threshold $t$.
\end{itemize}  
The result summary is a list of phrases. The shortest phrases contain $l$ words ($l$ is the window size of the convolutional neural layer). The longest phrases are the sentences themselves.

\section{Document Encoding and Relevance Modeling}
\subsection{Lexical Features}\label{sec:lex-matching}
We estimate the lexical features by performing lexical matching between a query and a  candidate case in different types of n-grams, skip-grams, longest common subsequence to measure various degrees of lexical similarity.

\begin{itemize}
    \item N-gram matching: measuring n-gram overlapping between a query and a candidate case. We employ unigram and bigram models.
    \item Skip-bigram matching: measuring the co-occurrence of all word pairs in their sentence order. This allows the same non-continuous word pairs could be found in both query and candidate.
    \item We also employ the unigram+skip-gram model which balances the unigram matching and skip-gram matching.
    \item Longest common subsequence: measuring the strictly ordered overlapping scattering over the texts. We employ two variants: standard version and distance-weighted version. The distance-weighted version favors subsequences with shorter distances among words. 
\end{itemize}

For each matching formula, we compute the matching scores by 3 different factors:
\begin{itemize}
    \item Recall: normalized by query, measuring the percentage of the query contents found in the candidate.
    \item Precision: normalized by candidate, measuring the percentage of the candidate contents found in the query.
    \item F-measure: harmony score of the previous two.
    $$f\text{-}measure=\frac{2 \times precision \times recall}{precision+recall}$$
\end{itemize}

To have more precise comparison between a query and a candidate, we apply the following 4 matching options: 
\begin{itemize}
    \item Summary vs. Summary: we compute the matching of the query's summary with the candidate's summary. This matching represents the comparison of the highlights between the query and the candidate.
    \item Paragraphs vs. Summary: we compute the matching of the query's paragraphs with the candidate's summary. This matching represents the ratio of the candidate summary mentioning relevant details.
    \item Summary vs. Paragraphs: we compute the matching of the query's summary with the candidate's paragraphs. This matching represents the ratio of the query's highlights mentioned in the candidate's details.
    \item Paragraphs vs. Paragraphs: we compute the matching of the query's paragraphs with the candidate's paragraphs. This matching represents the ratio of the query's details also occurred in the candidate's details.
\end{itemize}

For COLIEE 2019 dataset, since most of the candidate cases do not have a summary, we perform summary generation in two ways: using the lead sentence of each paragraph and the generated summary described in Section~\ref{sec:gen-text-sum}. This results in 6 matching options for COLIEE 2019 dataset.  

The coding for lexical features is in the form of q-c described as follows.
\begin{itemize}
    \item q is a subset of query components including its expert summary (s) and paragraphs (p).
    \item c is a subset of candidate components including its expert summary (s) and paragraphs (p). As the case of COLIEE 2019 dataset, we use the lead sentences (l) and the generated summary (e) instead of unavailable expert summary (s).
    \item Each component of q is compared with each component of c. 
\end{itemize}
For example, the lexical method sp-sp (q=sp, c=sp) means we perform 4 matching options: Summary vs. Summary, Summary vs. Paragraphs, Paragraphs vs. Summary, Paragraphs vs. Paragraphs, and the lexical method s-p (q=s, c=p) means we only perform Summary vs. Paragraphs matching. We use this naming for presenting lexical features' impact analysis in our experiments. 

In total, we collect lexical features from 6 matching formulas and 3 matching factors and 4 matching options, which results in 72 lexical features for measuring lexical matching between a query and each of its candidates. For COLIEE 2019 dataset, since most of the candidate cases do not have a summary, we perform summary generation in two ways: using the lead sentence of each paragraph and the generated summary described in Section~\ref{sec:gen-text-sum}. with the two additional matching options, we obtain 108 lexical features for COLIEE 2019 dataset.

\subsection{Latent Features in Continuous Vector Space}\label{sec:nn-features}
We utilize several approaches for encoding documents into continuous vector space as follows.
\begin{itemize}
    \item \textit{word-embeddings}: From word vectors, we apply three kinds of vector compositions for producing document vectors: max pooling, average pooling, hierarchical pooling. The word vectors can be obtained from word embedding models, for example, Google word2vec~\citep{mikolov2013distributed} or GloVe~\citep{pennington2014glove}. In this work, we use the pre-trained word embeddings published by Stanford University\footnote{Pre-trained with Wikipedia 2014 + Gigaword 5 (https://nlp.stanford.edu/projects/glove/)}. 
    Average pooling and max pooling are used to extract a fixed size feature vector from a sequence of vectors \citep{kim:2014:EMNLP2014,severyn2015learning,chen-etal-2017-enhanced}. Hierarchical pooling composes the document vector from the average pooling and the max pooling of the entire document, together with the average pooling of sentence level max pooling which adds sentence-boundary dependent features.

    \item \textit{doc2vec}\citep{le2014distributed}: This is a method for mapping text blocks into vector space. The method considers texts as sequences of tokens regardless of presented structures. 
    \item Encoded summarization: We apply our method described in Section \ref{sec:enc-sum}. The phrase scoring model is trained only on COLIEE 2018 dataset where the over 50K candidate cases have a summary. The pre-trained model is applied directly to COLIEE 2019 dataset without re-training.
To compare with this method, we also derive encoding methods based on the above \textit{word-embeddings}, and \textit{doc2vec} compositions, but apply them only on the expert summary part of each document. The derived methods are noted with ``(summary)".

\end{itemize}

\subsection{Query-Candidate Relevance Vector}\label{sec:rel-vec}
The relevance vector consists of the features indicating the relevance of a candidate given a query. We compose this vector from lexical features and latent features. 

The lexical features are computed by lexical matching which by themselves present the relevance measurement. 

For the latent features which are encoded information in continuous vector space, by comparing each dimension independently, we can estimate the compatibility of a query and a candidate over the dimension. Thus, we compute the relevance features from latent features as the element-wise product of query vector and candidate vector. First, we obtain query vector $\mathbf{g}(q)$ and candidate vector $\mathbf{g}(c)$ for each of the document vector compositions described in Subsection \ref{sec:nn-features}.
Then, we compute the relevance vector of query $q$ and candidate $c$ by the following element-wise product.
\begin{equation}
    \mathbf{h}(q,c) = \mathbf{g}(q) \odot \mathbf{g}(c)
\end{equation}

The combination of lexical features and latent features is presented in the query-candidate relevance vector as the concatenation of lexical matching features and the element-wise product of latent feature vectors of the query and the candidate.
\begin{equation}\label{eq:rel-vec}
    \mbox{relevance-vector}(q,c) = \left[ \mbox{lexical-features}(q,c) ; \mathbf{h}(q,c)\right]
\end{equation}

\section{Experiments}
\subsection{Summarization}
We trained the model with settings shown in Table \ref{tbl:sys-params}.  We use two sets of loss coefficients: (i) the parameters used for COLIEE 2018 submission, (ii) the parameters used for COLIEE 2019 submission. While the parameter set (i) is copied from~\citep{vutran2018mirel}, the parameter set (ii) is obtained by random searching around the set (i) for better retrieval performance on COLIEE 2018 dataset.

We report the empirical evaluation of the phrase scoring model applied to case summarization.
A predicted summary of a given case is composed according to Section~\ref{sec:gen-text-sum}. We evaluate the predicted summary with length-threshold $t$ values from 10\% to 50\% of document length. The evaluation is performed with ROUGE metrics including: ROUGE-1, ROUGE-2, ROUGE-SU. Results of the evaluation are shown in \ref{tbl:sum-results}. 

\begin{table}[H]
\caption{Phrase scoring model parameters. We use two sets of loss coefficients: (i) the parameters used for COLIEE 2018 submission, (ii) the parameters used for COLIEE 2019 submission. While the parameter set (i) is copied from~\citep{vutran2018mirel}, the parameter set (ii) is obtained by random searching around the set (i) for better retrieval performance on COLIEE 2018 dataset.}
\label{tbl:sys-params}
\centering
\begin{tabular}{l|l}
\hline
\hline
\textbf{Parameter} & \textbf{Description} \\
\hline
\hline
Embeddings (vector size $d$) & GloVe \citep{pennington2014glove}
$d=300$ \footnotemark  \\
CNN filters $c$ & 300 \\
CNN window size $l$ & 5 \\
MLP hidden size & 300 \\
Optimizer & Adam\citep{duchi2011adaptive} \\
Learning rate & 0.0001 \\
Gradient clipping max norm & 5.0 \\
Loss coefficients
$(a_1,a_2,b_1,b_2,b_3,b_4)$ & (i) $(1.0,1.0,0.5,0.1,0.01,0.02)$ \\
& (ii) $(1.0,1.7,0.3,0.7,0,0)$ \\
Size of negative set $|\{d'\}|$ & 2 \\
\hline
\hline

\end{tabular}
\end{table}
\footnotetext{Pre-trained with Wikipedia 2014 + Gigaword 5 (https://nlp.stanford.edu/projects/glove/)}

\begin{table}[H]
\centering
\caption{Summarization performance measured in ROUGE scores on dataset from COLIEE 2018 case law retrieval task. The phrase scoring model is trained with loss coefficients (i).}
\label{tbl:sum-results}
\begin{tabular}{|p{0.1\textwidth} | c | c | c | c | c  | c | c | c | c |}
\hline
\textbf{Length Threshold} $t$ & \multicolumn{3}{c|}{\textbf{ROUGE-1}} & \multicolumn{3}{c|}{\textbf{ROUGE-2}} & \multicolumn{3}{c|}{\textbf{ROUGE-SU6}} \\
& \textbf{Pre} & \textbf{Rec} & \textbf{F1} & \textbf{Pre} & \textbf{Rec} & \textbf{F1} & \textbf{Pre} & \textbf{Rec} & \textbf{F1} \\
\hline
10\%	&	0.482	&	0.409	&	0.405	&	0.186	&	0.152	&	0.152	&	0.258	&	0.199	&	0.167	\\
20\%	&	0.377	&	0.592	&	0.424	&	0.155	&	0.244	&	0.174	&	0.169	&	0.388	&	0.184	\\
30\%	&	0.304	&	0.687	&	0.390	&	0.135	&	0.311	&	0.174	&	0.116	&	0.511	&	0.155	\\
40\%	&	0.253	&	0.745	&	0.352	&	0.121	&	0.364	&	0.169	&	0.084	&	0.592	&	0.125	\\
50\%	&	0.216	&	0.784	&	0.318	&	0.109	&	0.407	&	0.162	&	0.063	&	0.651	&	0.100	\\

\hline
\end{tabular}
\end{table}

\begin{figure}
    \centering
    \includegraphics[width=0.7\textwidth]{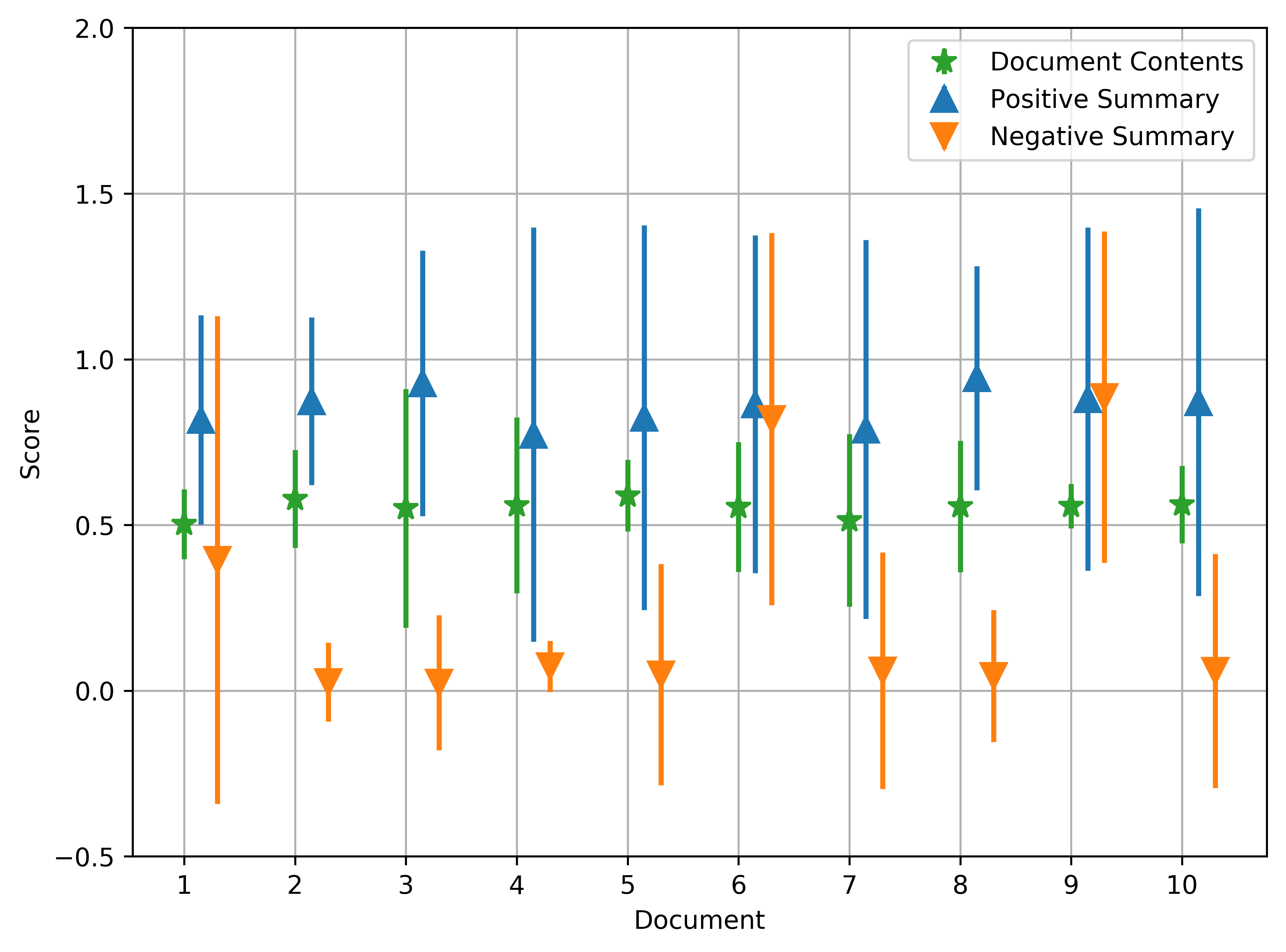}
    \caption{Visualization of score distribution (95\% confidence) per document showing the comparison among scores of a document's contents with its summary (positive summary) and other random document's summary (negative summary). Most of the sample cases, positive summaries have higher mean scores than document contents, and document contents have higher mean scores than negative summaries. }
    \label{fig:vs-score-perdoc}
\end{figure}

\begin{figure}
    \centering
    \includegraphics[width=0.6\textwidth]{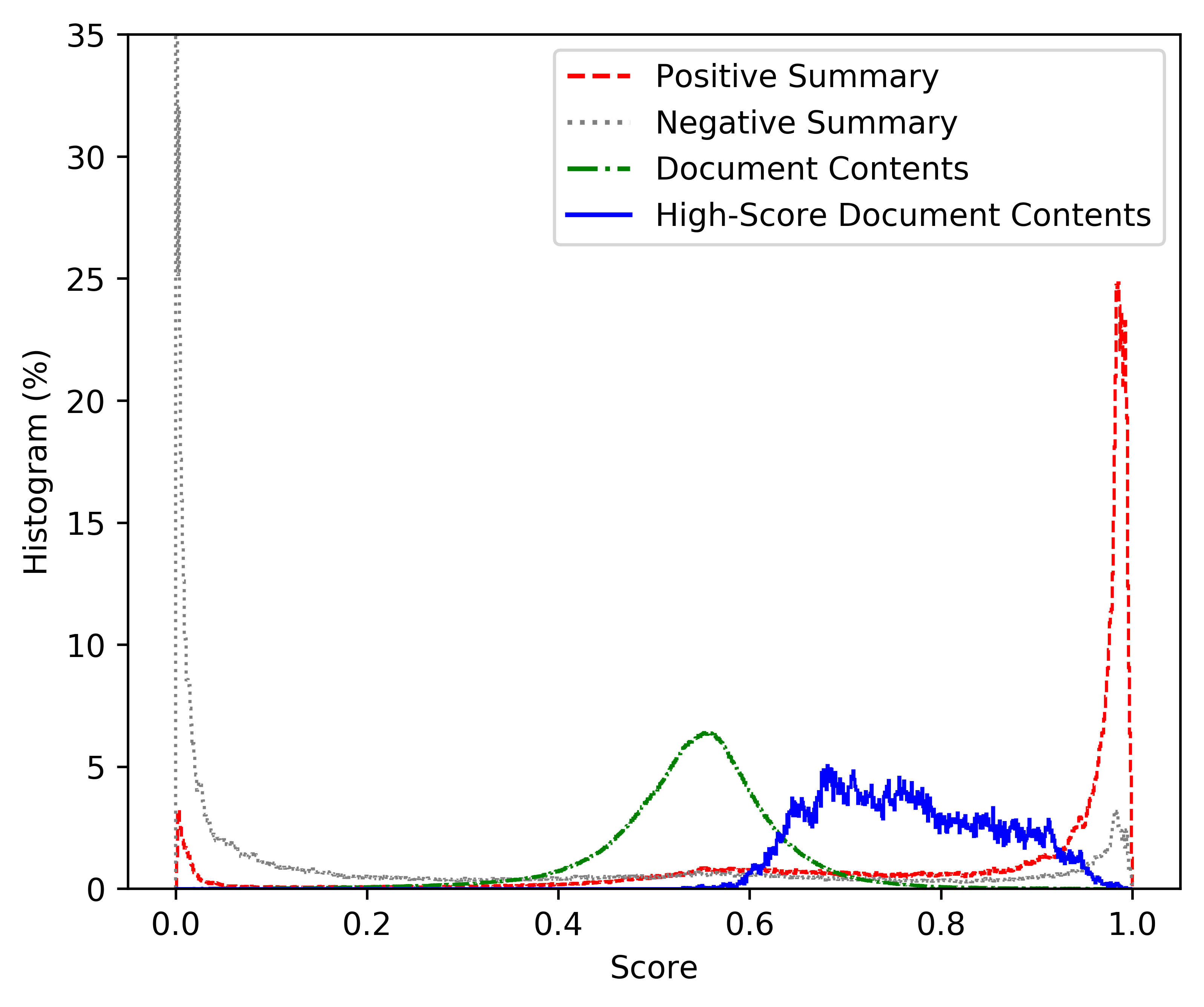}
    \caption{Visualization of score distribution over all data. Positive summaries have higher mean scores than document contents, and document contents have higher mean scores than negative summaries. High-score document contents are selected from top 50 highest phrases for each document. The phrases in high-score document contents affects much to the composition of document vectors, and could also be selected for summarizing documents.}
    \label{fig:vs-score-all}
\end{figure}

The phrase score statistics are shown in Fig.~\ref{fig:vs-score-perdoc} and Fig.~\ref{fig:vs-score-all}. 
Most of the sample cases, positive summaries have higher mean scores than document contents, and document contents have higher mean scores than negative summaries. As shown in Table~\ref{tbl:system-text-summary-example}, using our phrase scoring model, we can extract phrases similar to gold summary phrases.

We measure the score distribution but over all data. Similar to per document, positive summaries have higher mean scores than document contents, and document contents have higher mean scores than negative summaries. High-score document contents are selected from top 50 highest phrases for each document. The phrases in high-score document contents affects much to the composition of document vectors, and could also be selected for summarizing documents.

\begin{table}[H]
    \centering
    \begin{tabular}{l|p{7cm}}
        \textbf{Gold Summary} & Chan sought judicial review of a visa officer's decision denying his permanent residence application. He sought an injunction requiring the Minister to allow him to continue working and to take no enforcement action against him until his judicial review application was finally determined. The Federal Court of Canada, Trial Division, dismissed the motion. 
  
           \\ \hline
        \textbf{High-score Phrases} &  
- denying him an immigrant visa \par
- injunction requiring the Minister to \par
- for an immigrant visa moot \par
- judicial review of a visa \par
- existing application for an immigrant \par
- officer 's decision denying him \par
- application for an immigrant visa \par
- visa officer 's decision denying \par
- decision denying him an immigrant \par
- immigrant visa has been finally \\ \hline

        \textbf{Medium-score Phrases} & 
- Sciences ( Economics ) from \par
- This will be a matter \par
- might be good reasons that \par
- reasons that could bring s. \par
- of factors listed in column \par
- current employment authorization  . \par
- will arise for the applicant \par
- applied for permanent residence in \par
- his application issued on September \par
- its application , even though \par
        
        \\ \hline
        \textbf{Low-score Phrases} & 
- harm will arise as a \par
- turn then to the question \par
- While the applicant has demonstrated \par
- Immigration Regulations are : `` \par
- the applicant has demonstrated a \par
- established that irreparable harm will \par
- I turn then to the \par
- of the non-issuance of an \par
- he has not established that \par
- a serious question to be \par
        
        \\ \hline
    \end{tabular}
    \caption{Example outputs of phrase scoring model.}
    \label{tbl:system-text-summary-example}
\end{table}

For comparison, we evaluate on the dataset provided by \cite{hachey2004rhetorical}, which is a collection of 47 judgments of the House of Lord\footnote{https://www.parliament.uk/business/lords/} (HOLJ) from 2001 to 2003. We compare the results of sentence selection with the methods of \cite{hachey2004rhetorical} and \cite{kim-mi-young-legal-sum-10.1007/978-3-642-39931-2_14}. Since HOLJ corpus has only 47 documents, the phrase scoring model is trained on COLIEE 2018 dataset.  For the task of sentence selection, given a document $d = \{s\}$, we select top $t$ sentences with highest scores computed by sum of sentence (n-gram) phrase scores.  

\begin{itemize}
    \item \cite{hachey2004rhetorical}: develop a sentence classification method using models trained on several labor linguistic features: cue phrase, location, entities, sentence length, quotations, and thematic words.
    \item \cite{kim-mi-young-legal-sum-10.1007/978-3-642-39931-2_14}: develop a graph-based algorithm which selects sentences towards the conclusion/decision of the case. The sentences are connected based on the embedding probability, the probability that a sentence is embedded in another.
\end{itemize}

\begin{table}[H]
    \centering
    \caption{Sentence selection results by selection F-score on HOLJ corpus.}
    \label{tbl:sum-results-sent-sel-acc}
\begin{tabular}{|l | c | c | c |}
\hline
\textbf{Top} $t$ \textbf{Sentences} & \textbf{Pre} & \textbf{Rec} & \textbf{F1}  \\
\hline
10\%	&	0.197	&	0.136	&	0.155	\\
20\%	&	0.182	&	0.245   &   0.201	\\
30\%	&	0.168	&	0.344	&	0.219	\\
40\%	&	0.168	&	0.460	&	0.240	\\
50\%	&	0.171	&	0.579	&	0.258	\\
\hline \hline
Hachey et al. & 0.317 & 0.307 & 0.312 \\
Kim et al. & 0.313 & 0.364 & 0.337 \\
\hline
\end{tabular}
\end{table}

\begin{table}[H]
    \centering    
    \caption{Sentence selection results by ROUGE scores on HOLJ corpus.}
    \label{tbl:sum-results-sent-sel-rouge}
\begin{tabular}{|p{0.1\textwidth} | c | c | c | c | c  | c | c | c | c |}
\hline
\textbf{Top} $t$ \textbf{Sentences} & \multicolumn{3}{c|}{\textbf{ROUGE-1}} & \multicolumn{3}{c|}{\textbf{ROUGE-2}} & \multicolumn{3}{c|}{\textbf{ROUGE-SU6}} \\
& \textbf{Pre} & \textbf{Rec} & \textbf{F1} & \textbf{Pre} & \textbf{Rec} & \textbf{F1} & \textbf{Pre} & \textbf{Rec} & \textbf{F1} \\
\hline
10\%	&	0.523	&	0.715	&	0.583	&	0.313	&	0.424	&	0.347	&	0.302	&	0.530	&	0.342	\\
20\%	&	0.365	&	0.846	&	0.494	&	0.258	&	0.592	&	0.348	&	0.155	&	0.739	&	0.236	\\
30\%	&	0.289	&	0.896	&	0.424	&	0.221	&	0.685	&	0.325	&	0.097	&	0.824	&	0.164	\\
40\%	&	0.247	&	0.931	&	0.380	&	0.205	&	0.770	&	0.315	&	0.071	&	0.882	&	0.126	\\
50\%	&	0.220	&	0.957	&	0.350	&	0.194	&	0.838	&	0.307	&	0.056	&	0.928	&	0.103	\\

\hline
\end{tabular}

\end{table}

Even though, using the phrase scoring model, we can select sentences with high overlap with the gold sentences (Table~\ref{tbl:sum-results-sent-sel-rouge}), the accuracy of selecting the labeled sentences is low (Table~\ref{tbl:sum-results-sent-sel-acc}).
The results are understood as our phrase scoring model focuses on evaluating the importance of phrases, and is not directly learned to score sentences. 
Besides, there are two factors our phrase scoring model does not have during inference: (1) any explicit linguistic features other than word embedding, (2)  statistical information: term frequency-inverse document frequency. Furthermore, the phrase scoring model is trained on a different corpus. The common of the training corpus (COLIEE 2018) with the test corpus (HOLJ) is essentially captured through the use of word embedding.

\subsection{Retrieval}

In the data used in our experiments, the legal cases are sampled from a database of predominantly Federal Court of Canada case laws, provided by Compass Law. The data are provided by COLIEE competition~\citep{COLIEE2018} held in two years 2018 and 2019. 
In each of both the datasets, the data contain 285 queries, each query is attached with 200 candidate cases. 
Each candidate case is presented as a raw text document file which describes the details of the case. While a summary is presented in the query case, the candidate cases may not have summary section. 

We formulate the task as bipartite ranking problem and devise the learning to ranking method to solve it. We utilize pair-wise ranking strategy: pairing each noticed case with an irrelevant case from the candidate list. We adopt Linear-SVM as the learning algorithm for solving the optimization problem. The input of the learning-to-rank algorithm is the query-candidate relevance vectors obtained from Equation \ref{eq:rel-vec} in Section \ref{sec:rel-vec}. After obtaining the scored candidates as a ranked list, we proceed to select top $k$ highest scored candidates as the predicted noticed cases. 

The phrase scoring model was trained on only COLIEE 2018 dataset, and then adopted to generate encoded summarization vectors for case documents, and text summaries for the candidate cases in COLIEE 2019 dataset. For generating the text summaries, the summary length threshold $t$ (Section~\ref{sec:gen-text-sum}) is set to $t = 20\%$ document\text{-}length. As shown in Table~\ref{tbl:data-stat}, the average length of summaries is $\approx{10\%} $ document\text{-}length for COLIEE 2018 dataset, and $\approx{9\%}$ document\text{-}length for COLIEE 2019 dataset. Thus, with a threshold $t=20\% $ document\text{-}length, we could expect to cover potential information with good recall rate ($\approx 70\%$) while keeping an acceptable summary length.

\begin{table}
\centering
\caption{Validation results on COLIEE 2018 dataset. We select top 10 highest scored candidates when measuring precision, recall and f-measure. ``(summary)" indicates that  the corresponding encoding method is applied only on the summary part of the document.}
\label{tbl:val-results-2018}
\begin{tabular}{|p{6.8cm}|c|c|c|c|}
\hline \hline 
\textbf{Model}                                & \textbf{MAP} & \textbf{P} & \textbf{R} & \textbf{F1} \\ \hline \hline 
Lexical                                       & 0.530        & 0.420         & 0.520         & 0.398          \\ \hline
\textit{WordEmb}-Avg-pooling                          & 0.452        & 0.386         & 0.440         & 0.356          \\ 
\textit{WordEmb}-Max-pooling                          & 0.325        & 0.306        & 0.326        & 0.275          \\ 
\textit{WordEmb}-Hierarchical-pooling                          & 0.528        & 0.434         & 0.481         & 0.400          \\ 
\textit{doc2vec}                                       & 0.552        & 0.438         & 0.533         & 0.415          \\ \hline
\textit{WordEmb}-Avg-pooling (summary)           & 0.515        & 0.444         & 0.499         & 0.410          \\ 
\textit{WordEmb}-Max-pooling (summary)           & 0.400        & 0.370         & 0.362         & 0.324          \\ 
\textit{WordEmb}-Hierarchical-pooling (summary)           & 0.619        & 0.503         & 0.570         & 0.469          \\ 
\textit{doc2vec} (summary)                        & 0.422        & 0.367         & 0.407         & 0.334          \\ \hline
{\EncSum (i)}                               & {0.659}        & {0.510}         & {0.584}         & {0.478}          \\ 

\textbf{\EncSum (ii)}                               & \textbf{0.690}        & \textbf{0.529}         & \textbf{0.608}         & \textbf{0.494}          \\ \hline
 \hline 
Lexical+\textit{WordEmb}-Avg-pooling                & 0.686        & 0.522         & 0.653         & 0.502          \\ 
Lexical+\textit{WordEmb}-Max-pooling                & 0.687        & 0.515         & 0.642         & 0.494          \\ 
Lexical+\textit{WordEmb}-Hierarchical-pooling                & 0.772        & 0.565         & 0.705         & 0.545          \\ 
Lexical+\textit{doc2vec}                             & 0.684        & 0.518         & 0.644         & 0.496          \\ \hline
Lexical+\textit{WordEmb}-Avg-pooling (summary) & 0.688        & 0.528         & 0.646         & 0.505          \\ 
Lexical+\textit{WordEmb}-Max-pooling (summary) & 0.711        & 0.544         & 0.677         & 0.524          \\ 
Lexical+\textit{WordEmb}-Hierarchical-pooling (summary) & 0.783        & 0.579         & 0.725         & 0.560          \\ 
Lexical+\textit{doc2vec}(summary)              & 0.704        & 0.539         & 0.675         & 0.520          \\ \hline
{Lexical+\EncSum (i) }                     & {0.849}        & {0.601}         & {0.761}         & {0.583}          \\  

{Lexical+\EncSum (ii)}                     & \textbf{0.888}        & \textbf{0.623}         & \textbf{0.788}         & \textbf{0.607}          \\ \hline \hline 
\end{tabular}
\end{table}
\begin{table}
    \centering
    \caption{Lexical feature impact analysis by validation results on COLIEE 2018 dataset.  We select top 10 highest scored candidates when measuring precision, recall and f-measure. The coding for lexical features is in the form of q-c, where q is a subset of query components including a summary (s) and paragraphs (p), c is a subset of candidate components including a summary (s) and paragraphs (p). For example, the lexical method sp-ple (q=sp, c=sp) means we perform all 4 matching options, and the lexical method s-p (q=s, c=p) means we only compare the summary of a query with the paragraphs of a candidate.}
    \begin{tabular}{|p{6.8cm}|c|c|c|c|}
    \hline \hline
\textbf{Lexical Combination}		&	\textbf{MAP}	&	\textbf{P}	&	\textbf{R}	&	\textbf{F1}	\\ \hline \hline
s-s	&	0.372	&	0.331	&	0.378	&	0.302 \\ 
s-p	&	0.482	&	0.386	&	0.486	&	0.367 \\ 
p-s	&	0.435	&	0.356	&	0.434	&	0.331 \\ 
p-p	&	0.469	&	0.372	&	0.463	&	0.355 \\ [-0.75em] 
& & & &\\
sp-s	&	0.458	&	0.371	&	0.45	&	0.346 \\ 
sp-p	&	0.510	&	0.403	&	0.506	&	0.384 \\ [-0.75em]
& & & &\\
sp-sp	&   \textbf{0.530}   &   \textbf{0.420}   &   \textbf{0.520}   &   \textbf{0.398}   \\[-0.75em]
& & & & \\ 
\hline \hline
\end{tabular}
\label{tbl:lex-val-results-2018}
\end{table}

\begin{table}
    \centering
    \caption{Validation results on COLIEE 2019 dataset.  We select top 5 highest scored candidates when measuring precision, recall and f-measure.}
    \begin{tabular}{|p{6.8cm}|c|c|c|c|}
    \hline \hline
\textbf{Model}		&	\textbf{MAP}	&	\textbf{P}	&	\textbf{R}	&	\textbf{F1}	\\ \hline \hline
Lexical	&	0.715	&	0.495	&	0.641	&	0.485	\\
\hline 
\textit{WordEmb}-Avg-pooling & 0.218 & 0.177 & 0.210 & 0.161 \\ 
\textit{WordEmb}-Max-pooling & 0.270 & 0.223 & 0.260 & 0.206 \\ 
\textit{WordEmb}-Hierarchical-pooling & 0.417 & 0.331 & 0.405 & 0.311 \\
\textit{doc2vec} & 0.567 & 0.404 & 0.540 & 0.398 \\ \hline
\EncSum (i) & 0.542 & 0.430 & 0.516 & 0.402 \\ 
\EncSum (ii) & 0.576 & 0.436 & 0.534 & 0.410 \\ 
\hline \hline

Lexical+\textit{WordEmb}-Avg-pooling & 0.733 & 0.508 & 0.658 & 0.496 \\ 
Lexical+\textit{WordEmb}-Max-pooling & 0.750 & 0.526 & 0.679 & 0.513 \\ 
Lexical+\textit{WordEmb}-Hierarchical-pooling & 0.782 & 0.549 & 0.704 & 0.534 \\ 
Lexical+\textit{doc2vec} & 0.725 & 0.493 & 0.638 & 0.482 \\ 
\hline
Lexical+\EncSum (i) & 0.792 & 0.552 & 0.700 & 0.533 \\ 

Lexical+\EncSum (ii)&	\textbf{0.833}	&	\textbf{0.579}	&	\textbf{0.724}	&	\textbf{0.557}	\\
\hline \hline
    \end{tabular}
    \label{tbl:val-results-2019}
\end{table}

\begin{table}
    \centering
    \caption{Lexical feature impact analysis by validation results on COLIEE 2019 dataset.  We select top 5 highest scored candidates when measuring precision, recall and f-measure. The coding for lexical features is in the form of q-c, where q is a subset of query components including summary (s) and paragraphs (p), c is a subset of candidate components including paragraphs (p), lead sentences (l), and generated summary (e) (described in Section~\ref{sec:gen-text-sum}). For example, the lexical method sp-ple (q=sp, c=ple) means we perform all 6 matching options, and the lexical method s-p (q=s, c=p) means we only compare the summary of a query with the paragraphs of a candidate.}
    \begin{tabular}{|p{6.8cm}|c|c|c|c|}
    \hline \hline
\textbf{Lexical Combination}		&	\textbf{MAP}	&	\textbf{P}	&	\textbf{R}	&	\textbf{F1}	\\ \hline \hline
s-p		&	0.690	&	0.484	&	0.620	&	0.470	\\
s-l		&	0.589	&	0.420	&	0.528	&	0.405	\\
s-e		&	0.561	&	0.401	&	0.517	&	0.390	\\
p-p		&	0.680	&	0.476	&	0.601	&	0.461	\\
p-l		&	0.619	&	0.443	&	0.563	&	0.429	\\
p-e		&	0.588	&	0.413	&	0.534	&	0.402	\\[-0.75em] 
& & & &\\
sp-p	&	0.712	&	0.490	&	0.635	&	0.480	\\
sp-l	&	0.634	&	0.448	&	0.570	&	0.435	\\
sp-e	&	0.602	&	0.429	&	0.553	&	0.416	\\[-0.75em]
& & & &\\
sp-pl	&	0.713	&	0.493	&	0.639	&	0.483	\\
sp-pe	&	0.709	&	0.485	&	0.633	&	0.476	\\
sp-ple	&	\textbf{0.715}	&	\textbf{0.495}	&	\textbf{0.641}	&	\textbf{0.485}	\\[-0.75em]
& & & & \\ 
\hline \hline
\end{tabular}
\label{tbl:lex-val-results-2019}
\end{table}

We evaluated our approach by performing leave-one-out validation where we tested on each and every query from the provided 285 queries and the rest as training data.

We reported our system's validation results with the following metrics:
\begin{itemize}
    \item MAP: Mean average precision. 
    \item P, R, F1: Precision, Recall, F-measure whose values are averaged by query. This is straightforward as we average the results of all folds in the leave-one-out validation.
 \end{itemize}

The results in Tables \ref{tbl:val-results-2018}, \ref{tbl:val-results-2019}, \ref{tbl:test-results-2018}, \ref{tbl:test-results-2019}, \ref{tbl:competition-results-2018}, and \ref{tbl:competition-results-2019} show that \textbf{Lexical+\EncSum}, the combination of lexical features with encoded summarization, achieves the best performance.

The validation results of COLIEE 2018 (Table \ref{tbl:val-results-2018}) and COLIEE 2019 (Table~\ref{tbl:val-results-2019}) show that lexical features and latent features complement each other really well. The highest performance with either lexical or latent features is lower than the lowest performance of the combination. The improvement by the combination hints the existence of important information captured by latent features but not captured by lexical features. 

\textit{WordEmb}-Hierarchical-pooling performs better than \textit{WordEmb}-Max-pooling and \textit{WordEmb}-Avg-pooling. The hierarchical pooling consists of \textit{WordEmb}-Max-pooling, \textit{WordEmb}-Avg-pooling features and further sentence-level pooling which regards the sentence information boundary. 

As shown in Table~\ref{tbl:val-results-2018}, when limiting the document to only the summary part than the whole content, most of the models using \textit{WordEmb} or \textit{doc2vec} perform better, except \textit{doc2vec} without lexical features. This suggests the important of summarization in legal case retrieval task. 

The suggestion strongly presents in the results of the models using encoded summarization. The models with encoded summarization features outperforms other latent feature generation candidates including \textit{WordEmb}, \textit{doc2vec} on either the summary part or the whole document. Furthermore, the improvement of the encoded summarization suggests that this feature type not only embeds the summary properties of the document but also carries selectively important information from the document content. 

The above points also suggest that the summary of a case contains important information but may not contain all relevant information for case retrieval. This is intuitively seen as that the whole case may discuss various legal points besides the main points. Since the encoded summarization weights the case content based on the summary which contains the main points of the case, the other various legal points which are potentially related to the main points may be captured. 
Hence, the selectively carried information by the encoded summarization could be the related points to the main points of the case.

\begin{table}
\centering
\caption{Results on test data of COLIEE 2018. We select top 10 highest scored candidates when measuring precision, recall and f-measure. ``(summary)" indicates that  the corresponding encoding method is applied only on the summary part of the document.}
\label{tbl:test-results-2018}
\begin{tabular}{|p{7.8cm}|c|c|c|}
\hline \hline \textbf{Model}                                                  & \textbf{P} & \textbf{R} & \textbf{F1} \\ \hline 
Lexical                                                                          & 0.458                          & 0.429                          & 0.443                           \\ \hline \hline
\textit{WordEmb}-Avg-pooling                                   & 0.417                          & 0.391                          & 0.404                           \\ 
\textit{WordEmb}-Max-pooling                                   & 0.331                          & 0.310                          & 0.320                           \\ 
\textit{WordEmb}-Hierarchical-pooling                          & 0.493                          & 0.463                          & 0.477                           \\ 
\textit{doc2vec}                                                & 0.466                          & 0.437                          & 0.451                           \\ \hline
\textit{WordEmb}-Avg-pooling (summary)                    & 0.490                          & 0.459                          & 0.474                           \\ 
\textit{WordEmb}-Max-pooling (summary)                    & 0.432                          & 0.405                          & 0.418                           \\ 
\textit{WordEmb}-Hierarchical-pooling (summary)           & 0.585                          & 0.548                          & 0.566                           \\ 
\textit{doc2vec}(summary)                                 & 0.444                          & 0.417                          & 0.430                           \\ \hline
{\EncSum (i)}                                                            & \textit{0.598}                              & \textit{0.561}                              & \textit{0.579}                               \\ 
{\EncSum (ii)}                                                            & \textbf{0.608}                              & \textbf{0.571}                              & \textbf{0.589}                               \\ 
\hline  \hline 
Lexical+\textit{WordEmb}-Avg-pooling                         & 0.569                          & 0.534                          & 0.551                           \\ 
Lexical+\textit{WordEmb}-Max-pooling                         & 0.566                          & 0.531                          & 0.548                           \\ 
Lexical+\textit{WordEmb}-Hierarchical-pooling                & 0.607                          & 0.569                          & 0.587                           \\ 
Lexical+\textit{doc2vec}                                      & 0.571                          & 0.536                          & 0.553                           \\ \hline
Lexical+\textit{WordEmb}-Avg-pooling (summary)          & 0.578                          & 0.542                          & 0.559                           \\ 
Lexical+\textit{WordEmb}-Max-pooling (summary)          & 0.598                          & 0.561                          & 0.579                           \\ 
Lexical+\textit{WordEmb}-Hierarchical-pooling (summary) & 0.637                          & 0.598                          & 0.617                           \\ 
Lexical+\textit{doc2vec} (summary)                       & 0.622                          & 0.583                          & 0.602                           \\ \hline
{Lexical+\EncSum (i)}                                                  & \textit{0.676}                          & \textit{0.634}                          & \textit{0.655}                           \\
{Lexical+\EncSum (ii)}                                                  & \textbf{0.690}                          & \textbf{0.647}                          & \textbf{0.668}                           \\ \hline \hline 
\end{tabular}
\end{table}

\begin{table}
\centering
\caption{Results on test data of COLIEE 2019. We select top 5 highest scored candidates when measuring precision, recall and f-measure.}
\label{tbl:test-results-2019}
\begin{tabular}{|p{7.8cm}|c|c|c|}
\hline \multicolumn{4}{|c|}{}  \\[-1em] \hline \textbf{Model} & \textbf{P} & \textbf{R} & \textbf{F1} \\ \hline \multicolumn{4}{|c|}{}  \\[-1em] \hline 
Lexical	 & 	0.485	 & 	0.448	 & 	0.466 \\ 
\textit{WordEmb}-Avg-pooling	 & 	0.157	 & 	0.145	 & 	0.151 \\ 
\textit{WordEmb}-Max-pooling	 & 	0.239	 & 	0.221	 & 	0.230 \\ 
\textit{WordEmb}-Hierarchical-pooling	 & 	0.334	 & 	0.309	 & 	0.321 \\ 
\textit{doc2vec}	 & 	0.403	 & 	0.373	 & 	0.387 \\ \hline
{\EncSum (i)}	 & 	0.413	 & 	0.382	 & 	0.397 \\ 
{\EncSum (ii)}	 & 	0.426	 & 	0.394	 & 	0.409 \\ 
\hline \multicolumn{4}{|c|}{}  \\[-1em] 
Lexical+\textit{WordEmb}-Avg-pooling	 & 	0.489	 & 	0.452	 & 	0.469 \\ 
Lexical+\textit{WordEmb}-Max-pooling	 & 	0.541	 & 	0.500	 & 	0.520 \\ 
Lexical+\textit{WordEmb}-Hierarchical-pooling	 & 	0.590	 & 	0.545	 & 	0.567 \\ 
Lexical+\textit{doc2vec}	 & 	0.475	 & 	0.439	 & 	0.457 \\ \hline
{Lexical+\EncSum (i)}	 & 	0.544	 & 	0.503	 & 	0.523 \\ 
{Lexical+\EncSum (ii)}	 & 	\textbf{0.600}	 & 	\textbf{0.555}	 & 	\textbf{0.576} \\ 
\hline \hline
\end{tabular}
\end{table}

\begin{table}
\centering
\caption{Participants' results on test data of COLIEE 2018. We participated in the competition under the name "JNLP". "JNLP-k=10" is our best system utilizing the combination of lexical and encoded summarization using the base parameters.}
\label{tbl:competition-results-2018}
\begin{tabular}{|p{8cm}|c|c|c|}
\hline \multicolumn{4}{|c|}{}  \\[-1em] \hline \textbf{Model} & \textbf{P} & \textbf{R} & \textbf{F1} \\ \hline \multicolumn{4}{|c|}{}  \\[-1em] \hline

HUKB1 & 	0.497 & 	0.308 & 	0.381 \\ \hline
HUKB2 & 	0.405 & 	0.304 & 	0.347 \\ \hline
{JNLP-r=2.5} & 	{0.546} & 	{0.655} & 	{0.596} \\ \hline
{JNLP-k=10} & 	{0.676} & 	{0.634} & 	\textbf{0.655} \\ \hline
Smartlaw & 	0.287 & 	0.431 & 	0.345 \\ \hline
UA & 	0.372 & 	0.323 & 	0.346 \\ \hline
UA-postproc & 	0.348 & 	0.404 & 	0.374 \\ \hline
UA-smote & 	0.354 & 	0.393 & 	0.372 \\ \hline
UBIRLED-1 & 	0.133 & 	0.623 & 	0.219 \\ \hline
UBIRLED-2 & 	0.196 & 	0.720 & 	0.308 \\ \hline
UBIRLED-3 & 	0.561 & 	0.102 & 	0.172 \\ \hline
UL & 	0.564 & 	0.302 & 	0.393 \\ \hline \hline 
\end{tabular}
\end{table}

\begin{table}
    \centering
    \caption{Participants' results on test data of COLIEE 2019. We participated in the competition under the name "JNLP".  "JNLP.task\_1.p" is our best system utilizing the combination of lexical and encoded summarization using the pre-trained phrase scoring model.}
    \begin{tabular}{|l|p{6.8cm}|c|c|c|}
    \hline \hline
\textbf{Team}	&	\textbf{Run name}	&	\textbf{P}	&	\textbf{R}	&	\textbf{F1}	\\ \hline \hline
CACJ	&	submit\_task1\_CACJ01	&	0.212	&	0.585	&	0.311	\\ \hline
CLArg	&	CLarg	&	0.927	&	0.306	&	0.460	\\ \hline
HUKB	&	task1.HUKB	&	0.702	&	0.400	&	0.510	\\ \hline
IITP	&	task1.IITPBM25	&	0.626	&	0.385	&	0.477	\\ \hline
IITP	&	task1.IITPd2v	&	0.465	&	0.346	&	0.397	\\ \hline
IITP	&	task1.IITPdocBM	&	0.637	&	0.388	&	0.482	\\ \hline
ILPS	&	BERT\_Score\_0.946	&	0.681	&	0.433	&	0.530	\\ \hline
ILPS	&	BERT\_Score\_0.96	&	0.819	&	0.342	&	0.483	\\ \hline
ILPS	&	BM25\_Rank\_6	&	0.467	&	0.518	&	0.491	\\ \hline
\textit{JNLP}	&	\textit{JNLP.task\_1.p}	&	{0.593}	&	{0.549}	&	\textit{0.570}	\\ \hline
\textbf{JNLP}	&	\textbf{JNLP.task\_1.pl}	&	{0.600}	&	{0.555}	&	\textbf{0.576}	\\ \hline
\textbf{JNLP}	&	\textbf{JNLP.task\_1.ple}	&	{0.600}	&	{0.555}	&	\textbf{0.576}	\\ \hline
UA	&	UA\_0.52	&	0.351	&	0.336	&	0.344	\\ \hline
UA	&	UA\_0.54	&	0.364	&	0.324	&	0.343	\\ \hline
UA	&	UA\_0.57	&	0.356	&	0.333	&	0.344	\\ \hline \hline
    \end{tabular}
    \label{tbl:competition-results-2019}
\end{table}

\begin{table}
    \caption{An output example of our system from COLIEE 2018 test data. ``/*~...~*/": omitted.}
    \centering
    \begin{tabular}{p{8cm}|c|p{1.5cm}}
\textbf{Query}		\\
\hline
AstraZeneca applied for judicial review of a decision by the Minister of Health to disclose certain information related to AstraZeneca's supplementary new drug submission for LOSEC tablets (omeprozole magnesium) for the treatment of dyspepsia. AstraZeneca argued that: (1) the decision was a nullity because it was made by a person who lacked the authority to make the decision; and (2) the information requested was exempt from disclosure pursuant to ss. 20(1)(b) and (c) of the Access to Information Act or, alternatively, because it was either irrelevant to the request or had previously been severed. /*~...~*/		\\
\hline 
\hline 
\textbf{Candidates}	& \textbf{Noticed} &	\textbf{Ranked by our system} \\
\hline 

The Minister of Health released certain records related to the applicant's new drug submission. /*~...~*/ &	NO &	1 \\ \hline
Cyanamid applied under s. 44 of the Access to Information Act to review the Minister's decision to disclose the product monographs of two drugs and certain severed documents relating to the new drug submission for one of the drugs.  /*~...~*/	& YES &	2 \\ \hline
 /*~...~*/ All the proceedings were brought under the provisions of the Patented Medicines (Notice of Compliance) Regulations and concerned a drug containing the medicine known as omeprazole.  /*~...~*/ &	NO &	3 \\ \hline
        
The Minister of National Health and Welfare refused to permit Apotex Inc. to add information to its New Drug Submissions.   /*~...~*/ &	NO &	4 \\ \hline
Allergan Inc. commenced an application under the Patented Medicines (Notice of Compliance) Regulations respecting the '691 patent.   /*~...~*/ &	NO &	5 \\ \hline
The Minister of Transport decided to disclose records which included information about City Express Airline, operated by Air Atonabee Ltd. Air Atonabee applied for judicial review of the decision under s. 44 of the Access to Information Act. /*~...~*/ & YES & 7 \\ \hline
Brookfield Lepage Johnson Controls Facility Management Services (BLJC) provided professional facility management services to property owners and tenants across Canada. /*~...~*/  & NO & 8 \\ \hline
Sandoz moved under ss. 6(5)(a) and 6(5)(b) of the Patented Medicines (Notice of Compliance) Regulations, SOR/1993-133, for the dismissal in part of Abbott Laboratories' prohibition application in respect of clarithromycin patents which included patent 2,387,361. /*~...~*/ & NO & 9 \\ \hline
The Canadian Tobacco Manufacturers Council et al. applied pursuant to s. 44 of the Access to Information Act (Can.) for an order prohibiting the Minister of National Revenue from disclosing third party information. /*~...~*/ & YES & 10 \\ \hline \hline
Matol Botanical International Inc. applied to have four decisions authorizing disclosure of information relating to its business reviewed and set aside under s. 44 of the Access to Information Act. /*~...~*/ & YES & 12 \\ \hline
    \end{tabular}

    \label{tbl:concrete-example}
\end{table}

The validation results (Tables~\ref{tbl:lex-val-results-2018}, and ~\ref{tbl:lex-val-results-2019}) of lexical features with various combinations (from the 4 matching options for COLIEE 2018 and 6 matching options for COLIEE 2019) described in Section~\ref{sec:lex-matching} show that the combination of lexical matching options does have positive effect to improve the performance on both COLIEE 2018 and COLIEE 2019 datasets. On one hand, it is meaningful to have expert summaries for lexical matching as in COLIEE 2018, and on the other hand, pseudo/generated summaries could also help boost retrieval performance in COLIEE 2019 where candidate summaries are not available.

The encoded summarization (\EncSum) approach alone achieves MAP of 0.576 and F1 of 0.410 on COLIEE 2019 dataset, lower performance than the best lexical combination. The effect is different from the observation in COLIEE 2018 dataset where the performance of encoded summarization (MAP of 0.690 and F1 of 0.494) is higher than lexical matching approach. Since the encoded summarization model is  trained on only COLIEE 2018 dataset, some summary phenomena in COLIEE 2019 dataset may not be well captured. 

The combination of encoded summarization and lexical features does improve performance. The improvement by the combination of show that, even though the encoded summarization may not perform well alone, it still provides useful information for identifying relevant cases. 

Since our system use similarity as features for predicting supporting relationship, it has the limitation when it comes to non-supporting (unnoticed) but highly similar cases. As shown in Table~\ref{tbl:concrete-example}, in the top 10, many retrieved cases are about medicines even though they are not noticed. Aside from that, our system does retrieve the noticed cases (ranked 2, 7, and 10) which are actually diverse in topics.

\section{Related Work}
Legal case retrieval or retrieval of prior cases is an important research topic for decades where approaches to solve the corresponding task involve performing linguistics analysis, logical analysis, common lexical matching, and distributed vector representation with both common and legal expertise knowledge\citep{bench2012history}. In \citep{jackson2003information}, they build a system called ``History Assistant"  which extracts rulings from court opinions and retrieves relevant prior cases from a citator database by combining partial parsing techniques with domain knowledge and discourse analysis to extract information from the free text of court opinions.
In \citep{10.1007/11552413_49}, they develop a knowledge representation model for the intelligent retrieval of legal cases involving decomposing issues into sub-issues, and categorizing factors into pro-claimant, pro-responder and neutral factors. In \citep{Saravanan2009}, they overcome the problem of keyword-based search due to synonymy and ambivalence of words by developing an ontological framework to enhance the user’s query and ensure efficient retrieval by enabling inferences based on domain knowledge. Other works related to building legal ontology are \citep{wyner2008ontology,wyner2012legal,getman2014crowdsourcing}. Aside of linguistics approaches which are expensive to develop because of the required expertise knowledge, other approaches utilizes the emerging effectiveness of neural networks for natural language processing with the pioneer method of mapping texts to continuous vector space\citep{mikolov2013distributed,le2014distributed}. In \citep{mandal2017measuring}, the authors measure legal document similarity considering structural information of the document including paragraphs, summary and utilizing various representation methods including lexical features: TF-IDF, and topic modeling, and distributed vector representational features: \textit{word2vec}, and \textit{doc2vec}. They, however, do not perform the combination of those features. As lexical features and representational features may potential embed different information since they are extracted by different methodologies, the combination of them is promising.

\section{Conclusion}
We have presented our approach for modeling document summary into continuous vector space. We showed that our approach has positive signs in building an effective legal case retrieval system.
The results show the importance of exploiting the summary for solving legal case retrieval task. Furthermore, 
the improvement by the encoded summarization suggests that this feature type not only embeds the summary properties of the given case but also carries selectively important information from the case content which could be potentially related legal points to the main points of the case.
Furthermore, the combination of lexical features and latent features generated with neural networks yields positive results for solving the legal case retrieval task. 
The experimental results show that lexical features and latent features complement each other. The highest performance with either lexical or latent features is lower than the lowest performance of the combination. The improvement of the combination hints the existence of latent features not captured by lexical approach. 
We have also showed that the phrase scoring model trained from COLIEE 2018 dataset can provide useful features for representing documents in COLIEE 2019 dataset. 
There are several directions for improving the performance of legal case retrieval systems. 
One is that we can use the documents having a summary in COLIEE 2019 dataset for fine-tuning the phrase scoring model. Besides, the lexical matching has not yet considered the statistical information of terms in the corpus, which can be modeled by term frequency-inverse document frequency for example. Including such information may improve the matching by recognizing the statistically typical words for each document. 

\begin{acknowledgements}
This work was supported by JST CREST Grant Number JPMJCR1513, Japan.
\end{acknowledgements}

\bibliographystyle{spbasic} 
\bibliography{ref}

\begin{thebibliography}{27}
\providecommand{\natexlab}[1]{#1}
\providecommand{\url}[1]{{#1}}
\providecommand{\urlprefix}{URL }
\expandafter\ifx\csname urlstyle\endcsname\relax
  \providecommand{\doi}[1]{DOI~\discretionary{}{}{}#1}\else
  \providecommand{\doi}{DOI~\discretionary{}{}{}\begingroup
  \urlstyle{rm}\Url}\fi
\providecommand{\eprint}[2][]{\url{#2}}

\bibitem[{Bench-Capon et~al.(2012)Bench-Capon, Araszkiewicz, Ashley, Atkinson,
  Bex, Borges, Bourcier, Bourgine, Conrad, Francesconi
  et~al.}]{bench2012history}
Bench-Capon T, Araszkiewicz M, Ashley K, Atkinson K, Bex F, Borges F, Bourcier
  D, Bourgine P, Conrad JG, Francesconi E, et~al. (2012) A history of ai and
  law in 50 papers: 25 years of the international conference on ai and law.
  Artificial Intelligence and Law 20(3):215--319

\bibitem[{Chen et~al.(2017)Chen, Zhu, Ling, Wei, Jiang, and
  Inkpen}]{chen-etal-2017-enhanced}
Chen Q, Zhu X, Ling ZH, Wei S, Jiang H, Inkpen D (2017) Enhanced {LSTM} for
  natural language inference. In: Proceedings of the 55th Annual Meeting of the
  Association for Computational Linguistics (Volume 1: Long Papers),
  Association for Computational Linguistics, Vancouver, Canada, pp 1657--1668,
  \doi{10.18653/v1/P17-1152},
  \urlprefix\url{https://www.aclweb.org/anthology/P17-1152}

\bibitem[{Duchi et~al.(2011)Duchi, Hazan, and Singer}]{duchi2011adaptive}
Duchi J, Hazan E, Singer Y (2011) Adaptive subgradient methods for online
  learning and stochastic optimization. Journal of Machine Learning Research
  12(Jul):2121--2159

\bibitem[{Galgani et~al.(2012{\natexlab{a}})Galgani, Compton, and
  Hoffmann}]{galgani2012citation}
Galgani F, Compton P, Hoffmann A (2012{\natexlab{a}}) Citation based
  summarisation of legal texts. In: Pacific Rim International Conference on
  Artificial Intelligence, Springer, pp 40--52

\bibitem[{Galgani et~al.(2012{\natexlab{b}})Galgani, Compton, and
  Hoffmann}]{Galgani:2012:TAG:2238696.2238737}
Galgani F, Compton P, Hoffmann A (2012{\natexlab{b}}) Towards automatic
  generation of catchphrases for legal case reports. In: Proceedings of the
  13th International Conference on Computational Linguistics and Intelligent
  Text Processing - Volume Part II, Springer-Verlag, Berlin, Heidelberg,
  CICLing'12, pp 414--425, \doi{10.1007/978-3-642-28601-8_35},
  \urlprefix\url{http://dx.doi.org/10.1007/978-3-642-28601-8_35}

\bibitem[{Getman and Karasiuk(2014)}]{getman2014crowdsourcing}
Getman AP, Karasiuk VV (2014) A crowdsourcing approach to building a legal
  ontology from text. Artificial intelligence and law 22(3):313--335

\bibitem[{Hachey and Grover(2004)}]{hachey2004rhetorical}
Hachey B, Grover C (2004) A rhetorical status classifier for legal text
  summarisation. Text Summarization Branches Out

\bibitem[{Jackson et~al.(2003)Jackson, Al-Kofahi, Tyrrell, and
  Vachher}]{jackson2003information}
Jackson P, Al-Kofahi K, Tyrrell A, Vachher A (2003) Information extraction from
  case law and retrieval of prior cases. Artificial Intelligence
  150(1-2):239--290

\bibitem[{Johnson and Zhang(2015)}]{johnson-zhang:2015:NAACL-HLT}
Johnson R, Zhang T (2015) Effective use of word order for text categorization
  with convolutional neural networks. In: Proceedings of the 2015 Conference of
  the North American Chapter of the Association for Computational Linguistics:
  Human Language Technologies, Association for Computational Linguistics,
  Denver, Colorado, pp 103--112,
  \urlprefix\url{http://www.aclweb.org/anthology/N15-1011}

\bibitem[{Kalchbrenner et~al.(2014)Kalchbrenner, Grefenstette, and
  Blunsom}]{kalchbrenner-grefenstette-blunsom:2014:P14-1}
Kalchbrenner N, Grefenstette E, Blunsom P (2014) A convolutional neural network
  for modelling sentences. In: Proceedings of the 52nd Annual Meeting of the
  Association for Computational Linguistics (Volume 1: Long Papers),
  Association for Computational Linguistics, Baltimore, Maryland, pp 655--665,
  \urlprefix\url{http://www.aclweb.org/anthology/P14-1062}

\bibitem[{Kano et~al.(2018)Kano, Kim, Yoshioka, Lu, Rabelo, Kiyota, Goebel, and
  Satoh}]{COLIEE2018}
Kano Y, Kim MY, Yoshioka M, Lu Y, Rabelo J, Kiyota N, Goebel R, Satoh K (2018)
  Coliee-2018: Evaluation of the competition on legal information extraction
  and entailment. Twelfth International Workshop on Juris-informatics
  (JURISIN), COLIEE

\bibitem[{Kim et~al.(2013)Kim, Xu, and
  Goebel}]{kim-mi-young-legal-sum-10.1007/978-3-642-39931-2_14}
Kim MY, Xu Y, Goebel R (2013) Summarization of legal texts with high cohesion
  and automatic compression rate. In: Motomura Y, Butler A, Bekki D (eds) New
  Frontiers in Artificial Intelligence, Springer Berlin Heidelberg, Berlin,
  Heidelberg, pp 190--204

\bibitem[{Kim(2014)}]{kim:2014:EMNLP2014}
Kim Y (2014) Convolutional neural networks for sentence classification. In:
  Proceedings of the 2014 Conference on Empirical Methods in Natural Language
  Processing (EMNLP), Association for Computational Linguistics, Doha, Qatar,
  pp 1746--1751, \urlprefix\url{http://www.aclweb.org/anthology/D14-1181}

\bibitem[{Le and Mikolov(2014)}]{le2014distributed}
Le Q, Mikolov T (2014) Distributed representations of sentences and documents.
  In: Proceedings of the 31st International Conference on Machine Learning
  (ICML-14), pp 1188--1196

\bibitem[{Levy and Goldberg(2014)}]{levy2014dependency}
Levy O, Goldberg Y (2014) Dependency-based word embeddings. In: ACL (2), pp
  302--308

\bibitem[{Liu and Zhang(2018)}]{doi:10.1162COLIr00312}
Liu Y, Zhang M (2018) Neural network methods for natural language processing.
  Computational Linguistics 44(1):193--195, \doi{10.1162/COLI\_r\_00312},
  \urlprefix\url{https://doi.org/10.1162/COLI_r_00312},
  \eprint{https://doi.org/10.1162/COLI_r_00312}

\bibitem[{Mandal et~al.(2017{\natexlab{a}})Mandal, Chaki, Saha, Ghosh, Pal, and
  Ghosh}]{mandal2017measuring}
Mandal A, Chaki R, Saha S, Ghosh K, Pal A, Ghosh S (2017{\natexlab{a}})
  Measuring similarity among legal court case documents. In: Proceedings of the
  10th Annual ACM India Compute Conference on ZZZ, ACM, pp 1--9

\bibitem[{Mandal et~al.(2017{\natexlab{b}})Mandal, Ghosh, Pal, and
  Ghosh}]{Mandal:2017:ACI:3132847.3133102}
Mandal A, Ghosh K, Pal A, Ghosh S (2017{\natexlab{b}}) Automatic catchphrase
  identification from legal court case documents. In: Proceedings of the 2017
  ACM on Conference on Information and Knowledge Management, ACM, New York, NY,
  USA, CIKM '17, pp 2187--2190, \doi{10.1145/3132847.3133102},
  \urlprefix\url{http://doi.acm.org/10.1145/3132847.3133102}

\bibitem[{Mikolov et~al.(2013)Mikolov, Sutskever, Chen, Corrado, and
  Dean}]{mikolov2013distributed}
Mikolov T, Sutskever I, Chen K, Corrado GS, Dean J (2013) Distributed
  representations of words and phrases and their compositionality. In: Advances
  in neural information processing systems, pp 3111--3119

\bibitem[{Olsson and of~Judicial~Administration(1999)}]{olsson1999guide}
Olsson L, of~Judicial~Administration AI (1999) Guide to Uniform Production of
  Judgments. Australian Institute of Judicial Administration,
  \urlprefix\url{https://books.google.co.jp/books?id=mKnAAQAACAAJ}

\bibitem[{Pennington et~al.(2014)Pennington, Socher, and
  Manning}]{pennington2014glove}
Pennington J, Socher R, Manning C (2014) Glove: Global vectors for word
  representation. In: Proceedings of the 2014 conference on empirical methods
  in natural language processing (EMNLP), pp 1532--1543

\bibitem[{Saravanan et~al.(2009)Saravanan, Ravindran, and
  Raman}]{Saravanan2009}
Saravanan M, Ravindran B, Raman S (2009) Improving legal information retrieval
  using an ontological framework. Artificial Intelligence and Law
  17(2):101--124, \doi{10.1007/s10506-009-9075-y},
  \urlprefix\url{https://doi.org/10.1007/s10506-009-9075-y}

\bibitem[{Severyn and Moschitti(2015)}]{severyn2015learning}
Severyn A, Moschitti A (2015) Learning to rank short text pairs with
  convolutional deep neural networks. In: Proceedings of the 38th International
  ACM SIGIR Conference on Research and Development in Information Retrieval,
  ACM, pp 373--382

\bibitem[{Tran et~al.(2018)Tran, Nguyen, and Satoh}]{vutran2018mirel}
Tran VD, Nguyen ML, Satoh K (2018) Automatic catchphrase extraction from legal
  case documents via scoring using deep neural networks. In: Workshop on MIning
  and REasoning with Legal texts

\bibitem[{Wyner(2008)}]{wyner2008ontology}
Wyner A (2008) An ontology in owl for legal case-based reasoning. Artificial
  Intelligence and Law 16(4):361

\bibitem[{Wyner and Hoekstra(2012)}]{wyner2012legal}
Wyner A, Hoekstra R (2012) A legal case owl ontology with an instantiation of
  popov v. hayashi. Artificial Intelligence and Law 20(1):83--107

\bibitem[{Zeng et~al.(2005)Zeng, Wang, Zeleznikow, and
  Kemp}]{10.1007/11552413_49}
Zeng Y, Wang R, Zeleznikow J, Kemp E (2005) Knowledge representation for the
  intelligent legal case retrieval. In: Khosla R, Howlett RJ, Jain LC (eds)
  Knowledge-Based Intelligent Information and Engineering Systems, Springer
  Berlin Heidelberg, Berlin, Heidelberg, pp 339--345

\end{thebibliography}
\end{document}